\documentclass[10pt,twocolumn,letterpaper]{article}

\usepackage{iccv}
\usepackage{times}
\usepackage{epsfig}
\usepackage{graphicx}
\usepackage{amsmath}
\usepackage{amssymb}
\usepackage{subcaption}

\usepackage[pagebackref=true,breaklinks=true,letterpaper=true,colorlinks,bookmarks=false]{hyperref}
\DeclareMathOperator{\E}{\mathbb{E}}
\iccvfinalcopy 

\def\iccvPaperID{6568} 

\usepackage{color}
\definecolor{light}{rgb}{0.4, 0.4, 0.4}
\def\light#1{{\color{light}#1}}
\ificcvfinal\fi
\begin{document}

\title{Uncertainty-aware Audiovisual Activity Recognition using Deep Bayesian Variational Inference}

\author{Mahesh Subedar\thanks{Contributed equally.}~,~~~Ranganath Krishnan\footnotemark[1]~,~~~Paulo Lopez Meyer,~~~Omesh Tickoo,~~~Jonathan Huang~~\\
\and
\light{\small\{mahesh.subedar, ranganath.krishnan, paulo.lopez.meyer, omesh.tickoo, jonathan.huang\}@intel.com}
\\
~~~~~~~~~~~~~~~~~~~~~~~~~~~~~~~~~Intel Labs~~~~~~~~~~~~~~~~~~~~~~~~~~~
}
\maketitle

\begin{abstract}
Deep neural networks (DNNs) provide state-of-the-art results for a multitude of applications, but the approaches using DNNs for multimodal audiovisual applications do not consider predictive uncertainty associated with individual modalities.
Bayesian deep learning methods provide principled confidence and quantify predictive uncertainty.
Our contribution in this work is to propose an uncertainty aware multimodal Bayesian fusion framework for activity recognition. We demonstrate a novel approach that combines deterministic and variational layers to scale Bayesian DNNs to deeper architectures. Our experiments using in- and out-of-distribution samples selected from a subset of Moments-in-Time (MiT) dataset show a more reliable confidence measure as compared to the non-Bayesian baseline and the Monte Carlo dropout (MC dropout) approximate Bayesian inference. We also demonstrate the uncertainty estimates obtained from the proposed framework can identify out-of-distribution data on the UCF101 and MiT datasets. In the multimodal setting, the proposed
framework improved precision-recall AUC by 10.2\% on the subset of MiT dataset as compared to non-Bayesian baseline.
\end{abstract}

\section{Introduction}
Vision and audio are complementary inputs and fusing these modalities can greatly benefit an activity recognition application.
Multimodal audiovisual activity recognition using deep neural network (DNN) architectures are not successful in modeling the inherent ambiguity in the correlation between two modalities.
One of the modalities (e.g., sneezing in audio, writing in vision) can be more certain about the activity class than the other modality. It is important to model reliable uncertainty estimates for the individual modalities to benefit from multimodal fusion.

DNNs trained on large datasets~\cite{kay2017kinetics,abu2016youtube,monfort2018moments} have been successful in solving many perception tasks with state-of-the-art results.
However, DNNs are trained to obtain the maximum likelihood estimates and disregard uncertainty around the model parameters that eventually can lead to predictive uncertainty. Deep learning models may fail in the case of noisy or out-of-distribution data, leading to overconfident decisions that could be erroneous as softmax probability does not capture overall model confidence. Instead, it represents relative probability that an input is from a particular class compared to the other classes.

Probabilistic Bayesian models provide principled ways to gain insight about data and capture reliable uncertainty estimates in predictions. Bayesian deep learning~\cite{neal2012bayesian,gal2016uncertainty} has allowed bridging DNNs and probabilistic Bayesian theory to leverage the strengths of both methodologies.  Bayesian deep learning framework with Monte Carlo (MC) dropout approximate inference \cite{gal2016dropout} is used in visual scene understanding applications including camera relocalization \cite{kendall2016modelling}, semantic segmentation \cite{kendall2015bayesian} and depth regression \cite{kendall2017multi}.

\begin{figure}[t]
\centering
\includegraphics[width=0.95\linewidth]{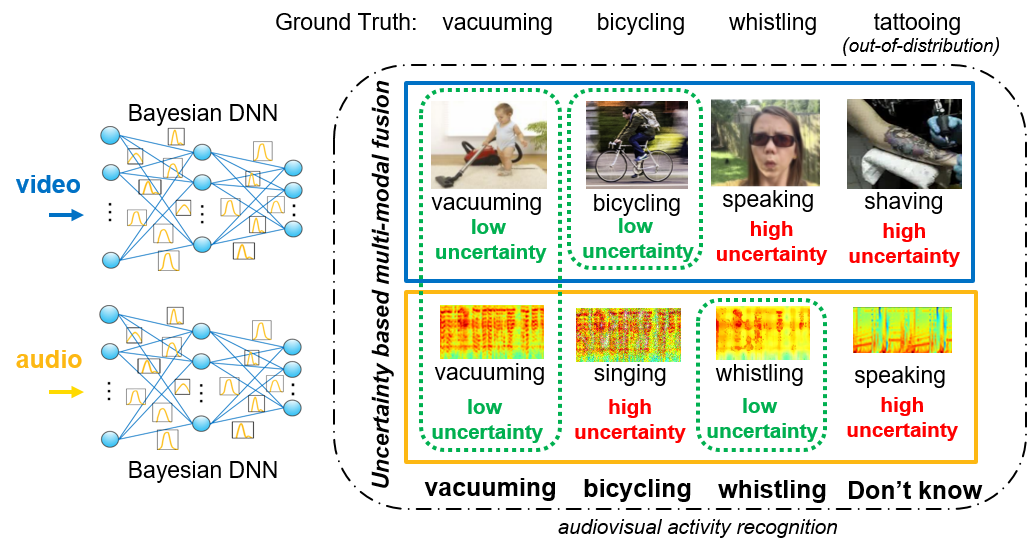}
\caption{\small  Uncertainty-aware audiovisual activity recognition}
\label{fig:high_blk_diag}
\end{figure}

Activity recognition is an active area of research with multiple approaches depending on the application domain and the types of sensors~\cite{lara2013survey}. Human activity recognition using wearable sensors such as accelerometer/gyroscopes and heart-rate monitors is used to recognize everyday human activities that include walking, running, and swimming. Human pose-based activity recognition~\cite{raptis2013poselet,sigurdsson2017actions} methods aggregate motion and appearance information along tracks of human body parts to recognize human activity. Multimodal methods which combine optical flow or depth information along with RGB data~\cite{simonyan2014two,shahroudy2018deep} are shown to provide state-of-the-art results for generic (not just human) activity recognition tasks. Methods which combine semantic level information~\cite{ziaeefard2015semantic} such as pose, object/scene context and other attributes including linguistic descriptors have been proposed to detect group activities.
~~~~~~~~~~~~~~~~~~~~~~~~~~~~~~~~~~~~~~~~~~~~~~~~~~~~~~~~~~~~~~~~~~~~~~~~~~~~~~~~~~~~~~~~~~~~~~~~

In this work, we focus on audiovisual activity recognition and use Bayesian DNN with stochastic variational inference~(VI) to reliably estimate uncertainty associated with the individual modalities for multimodal fusion (as shown in Figure~\ref{fig:high_blk_diag}). 

Our main contributions in this work include:
\begin{enumerate}
\item A multimodal fusion framework based on predictive uncertainty estimates applied to activity recognition: To the best of our knowledge, this is the first work on multimodal fusion based on uncertainty estimates using Bayesian deep learning with variational inference.
\item A scalable variational inference with hybrid Bayesian DNN architecture by combining deterministic and variational layers.
\item Identifying out-of-distribution data for audiovisual activity recognition using uncertainty estimates: We demonstrate the uncertainty estimates obtained from the proposed architecture can identify out-of-distribution data in Moments-in-Time (MiT) and UCF-101 action recognition datasets.
\end{enumerate}

The rest of the document is divided into the following sections. The background on Bayesian DNNs and audiovisual activity recognition are presented in Section~\ref{sec:background}. In Section~\ref{sec:architecture}, the proposed Bayesian multimodal DNN architecture is presented. The results are presented in Section~\ref{sec:results}, followed by  conclusions in Section~\ref{sec:conclusion}.

\section{Background}
\label{sec:background}

\subsection{Bayesian deep neural networks}
\begin{figure}[t]
\centering
\includegraphics[width=0.85\linewidth]{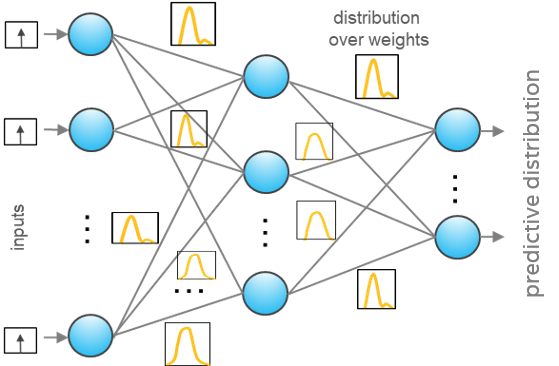}
\caption{ Bayesian neural network}
\label{fig:bnn}
\end{figure}

Bayesian DNNs provide a probabilistic interpretation of deep learning models by placing distributions over the model parameters (shown in Figure~\ref{fig:bnn}). Bayesian inference can be applied to estimate the predictive distribution by propagating over the model likelihood while marginalizing over the learned posterior parameter distribution. Bayesian DNNs also help in regularization by introducing distribution over network parameters, capturing the posterior uncertainty around the neural network parameters. This allows transferring inherent DNN uncertainty from the parameter space to the predictive uncertainty.

Given training dataset $D=\{x,y\}$ with inputs $x = {x_1, . . . , x_N}$ and their corresponding outputs $y = {y_1, . . . , y_N}$, in parametric Bayesian setting we would like to infer a distribution over parameters $w$ as a function $y = f_w(x)$ that represents the DNN model. With the posterior for model parameters inferred during Bayesian neural network training, we can predict the output for a new data point by propagating over the model likelihood $p(y|x,w)$ while drawing samples from the learned parameter posterior $p(w|D$).
Equation~\ref{eq:post_dist} shows the posterior distribution of model parameters obtained from model likelihood.
\begin{equation}
p(w|D) = \frac{p(y|x,w)p(w)}{p(y|x)}
\label{eq:post_dist}
\end{equation}

Computing the posterior distribution $p(w|D)$ is often intractable, some of the previously proposed techniques to achieve an analytically tractable inference include:~(i)~Markov Chain Monte Carlo (MCMC) sampling based probabilistic inference \cite{neal2012bayesian,welling2011bayesian} (ii)~variational inference techniques to infer the tractable approximate posterior distribution around model parameters \cite{graves2011practical,ranganath2013black,blundell2015weight} and (iii)~Monte Carlo dropout approximate inference \cite{gal2016dropout}. 

Variational inference \cite{jordan1999introduction, blei2017variational} is an active area of research in Bayesian deep learning, which uses gradient based optimization. 
This technique approximates a complex probability distribution $p(w|D)$ with a simpler distribution $q_\theta(w)$, parameterized by variational parameters $\theta$ while minimizing the Kullback-Leibler (KL) divergence~\cite{bishop2006pattern}. Minimizing the KL divergence is equivalent to maximizing the log evidence lower bound \cite{bishop2006pattern,gal2016dropout}.
\begin{equation}
\begin{aligned}
\mathcal{L} := \int q_\theta(w)\,log\,p(y|x,w)\,dw\\
~~~~~~~~~~~~~~~~~~~~ - KL[q_\theta(w)||p(w)]
\end{aligned}
\label{eq_ELBO}
\end{equation}

Predictive distribution is obtained through multiple stochastic forward passes through the network during the prediction phase while sampling from the posterior distribution of network parameters through Monte Carlo estimators.  Equation~\ref{eq:pred_dist} shows the predictive distribution of the output $y^*$ given new input $x^*$:
\begin{equation}
\begin{gathered}
p(y^*|x^*,D) = \int p(y^*|x^*,w)\,q_\theta(w) dw \\
p(y^*|x^*,D) \approx \frac{1}{T} \sum_{i=1}^{T}p(y^*|x^*,w_i)~,~~~w_i\sim q_\theta(w)
\end{gathered}
\label{eq:pred_dist}
\end{equation}
where, $T$ is number of Monte Carlo samples.

In~\cite{gal2016uncertainty, kendall2017multi}, modeling aleatoric and epistemic uncertainty is described.
We evaluate the epistemic uncertainty using Bayesian active learning by disagreement (BALD) \cite{houlsby2011bayesian} for the activity recognition task. BALD quantifies mutual information between parameter posterior distribution and predictive distribution, as shown in Equation~\ref{eq:mutual information}.
\begin{equation}
BALD := H(y^*|x^*, D)-\E_{p(w|D)}[H(y^*|x^*, w)]\\
\label{eq:mutual information}
\end{equation}
where, $H(y^*|x^*, D)$ is the predictive entropy which captures a combination of aleatoric and epistemic uncertainty given by:
\begin{equation}
H(y^*|x^*, D)=-\sum_{i=0}^{K-1}p_{i\mu}\,log\,p_{i\mu}\\
\label{eq:pred_entropy}
\end{equation}
$p_{i\mu}$ is predictive mean probability of $i^{th}$ class from $T$ Monte Carlo samples and K is the total number of output classes.
~~~~~~~~~~~~~~~~~~~~~~~~~~~~~~~~~~~~~~~~~~~~~~~~~~~~~~~~~~~~~~~~~~~~~~~~~~~~~~~

\subsection{Audiovisual Activity Recognition}

Vision and audio are the ubiquitous sensor inputs which are complementary in nature and have different representations. Audiovisual methods apply joint modeling of the audio and vision inputs~\cite{ngiam2011multimodal,afouras2018deep} to achieve higher accuracies for complex tasks such as activity recognition.

Multimodal models are proposed for audiovisual analysis tasks such as emotion recognition~\cite{srivastava2012learning}, audiovisual speech recognition~\cite{ngiam2011multimodal}, speech localization~\cite{gao2018learning,owens2018audio}, cross-modal retrieval~\cite{aytar2017see}.
The audiovisual speech recognition (AVSR) task is shown to benefit from multimodal training of the joint models. In~\cite{ngiam2011multimodal}, a deep autoencoder model for cross-modality feature learning is proposed, where better features for one modality can be learned if multiple modalities are present at training time. A deep audio-visual speech recognition model~\cite{afouras2018deep} using self-attention encoder architecture is proposed to recognize speech from talking faces using vision and audio inputs. Recent work on sound localization and separation~\cite{gao2018learning,owens2018audio} has shown the benefits of a joint audiovisual representation for cross-modal self-supervised learning using only audio-visual correspondence as the objective function. These audiovisual methods apply joint modeling of the audio and vision inputs during the training phase for better generalizability of the models, but then use single modality during the inference phase. None of the methods listed here provide a quantifiable means to determine the relative importance of each modality.

Vision-based activity recognition techniques apply a combination of spatiotemporal models~\cite{tran2018closer,arandjelovic2016netvlad,zhou2018temporal} to capture pixel-level information and temporal dynamics of the scene. In recent years, visual activity recognition models often use ConvNets-based models for spatial feature extraction. The image-based models~\cite{he2016deep,szegedy2015going} are pre-trained on ImageNet dataset to represent the spatial features. The temporal dynamics for activity recognition~\cite{wang2016temporal,zhou2017temporal} is typically modeled either by using a separate temporal sequence modeling such as variants of RNNs~\cite{donahue2015long,yuan2017temporal} or by applying 3D ConvNets~\cite{carreira2017quo}, which extend 2D ConvNets to the temporal dimension.  Bayesian neural network is used for visual activity recognition~\cite{krishnan2018bar} to capture uncertainty estimates.

Following the successes of ConvNets on vision tasks, they are shown to provide state-of-the-art results for audio classification as well. Many of the top performing methods from recent audio classification challenges~\cite{mesaros2018multi,piczak2015esc} use DNN architectures~\cite{sakashita2018acoustic,Dorfer2018,Zeinali2018} with convolutional layers. In~\cite{hershey2017cnn}, a  model similar to the VGG architecture (VGGish model) from the vision domain was trained using log-Mel spectrogram features on the Audio Set~\cite{gemmeke2017audio} dataset. Audio Set contains over one million Youtube video samples labeled with a vocabulary of acoustic events.

In this work, we focus on audiovisual activity recognition using Bayesian DNNs on the trimmed video samples. The 3D-ConvNet (C3D) architecture~\cite{tran2015learning} is shown to provide generic spatiotemporal representation for multiple vision tasks. We use a variant of 3D-ConvNet ResNet-101 C3D~\cite{hara3dcnns} architecture for the visual representation. We use VGGish architecture~\cite{hershey2017cnn} for audio representation, which is shown to provide generic features for audio classification tasks.

\begin{figure*}
\begin{center}
\includegraphics[width=0.85\linewidth]{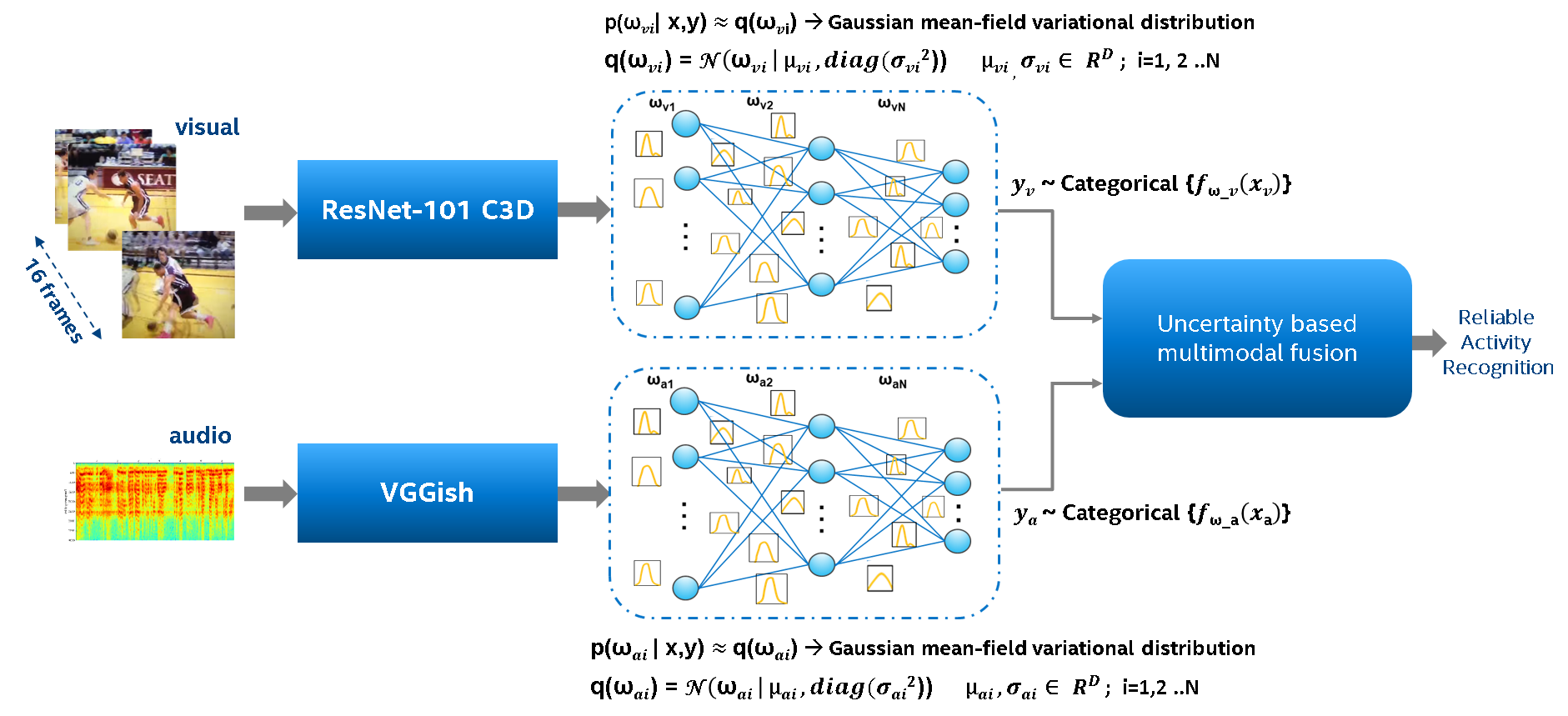}
\end{center}
\caption{ Bayesian audiovisual activity recognition: ResNet-101 C3D and VGGish DNN architectures are used to represent vision and audio information, respectively. The final layer of the DNN is replaced with three fully connected variational layers followed by categorical distribution. The Bayesian inference is applied to the variational layers through Monte Carlo sampling on the posterior of model parameters, which provides the predictive distribution.}
\label{fig:VisionBlkDiagram}
\end{figure*}
\section{Bayesian Multimodal DNN Architecture}
\label{sec:architecture}
We present a Bayesian multimodal fusion framework based on uncertainty estimates for audiovisual activity recognition.
The block diagram of the proposed audiovisual activity recognition using Bayesian variational inference is shown in Figure~\ref{fig:VisionBlkDiagram}. 
We use the ResNet-101 C3D and VGGish architectures for visual and audio modalities, respectively.
We replace the final fully connected layer for both vision and audio DNN models with three fully connected variational layers followed by the categorical distribution.

The weights in fully connected variational layers are modeled through mean-field Gaussian distribution, and the network is trained using Bayesian variational inference based on KL divergence~\cite{ranganath2013black,blundell2015weight}.
In order to learn the posterior distribution of model parameters $w$, we train Bayesian DNN with variational inference method. The objective is to optimize log evidence lower bound (ELBO) (Equation~\ref{eq_ELBO}) as the cost function.
The model parameters of the fully connected variational layers are parametrized by mean $\mu$ and variance $\sigma^2$, i.e. $q_\theta(w)=\mathcal{N}(w|\mu, \sigma^2)$.
These parameters in the variational layers are optimized by minimizing the negative ELBO loss~($L^v$)~\cite{bishop2006pattern}:
\begin{equation}
\begin{aligned}
L^v=-\mathbb{E}_{q_\theta(w)}[log\,p(y|x,w)]+KL[q_\theta(w)||p(w)]\\
\end{aligned}
\label{eq:ELBO_loss}
\end{equation}
$~~~~~~~\mu_{i+1}\leftarrow\mu_i-\alpha\,\Delta_{\mu}L^v_i~~~\,\,\,\,\,\,\,\,\,\,\,\,\sigma_{i+1}\leftarrow\sigma_i-\alpha\,\Delta_{\sigma}L^v_i\\\\$ where, $i$ is the training step, $\alpha$ is the learning rate, $\Delta_{\mu}L^v$ and $\Delta_{\sigma}L^v$ are gradients of the loss function computed w.r.t $\mu$ and $\sigma$, respectively.
We use Flipout~\cite{wen2018flipout}, which is an efficient method that decorrelates the gradients within a mini-batch by implicitly sampling pseudo-independent weight perturbations for each input.

The parameters in deterministic layers are optimized using cross-entropy loss ($L^d$)~\cite{goodfellow2016deep} given by: 
\begin{equation}
L^d=-\sum_c\,y_c\log\,\hat{y_c}
\label{eq:cross_entropy_eq}
\end{equation}
where, $y_c$ and $\hat{y_c}$ are true and predicted label distributions, respectively.
The model parameters for variational and deterministic DNN layers are obtained by applying stochastic gradient descent optimizer~\cite{bottou2010large} to the loss functions given in Equation~\ref{eq:ELBO_loss} and \ref{eq:cross_entropy_eq}, respectively.
During prediction stage we perform multiple Monte Carlo forward passes on the final variational layers by sampling the parameters from learned posteriors to measure uncertainty estimates using Equation~\ref{eq:mutual information}~\&~\ref{eq:pred_entropy}.

\begin{figure}[t]
\centering
\includegraphics[width=0.5\linewidth]{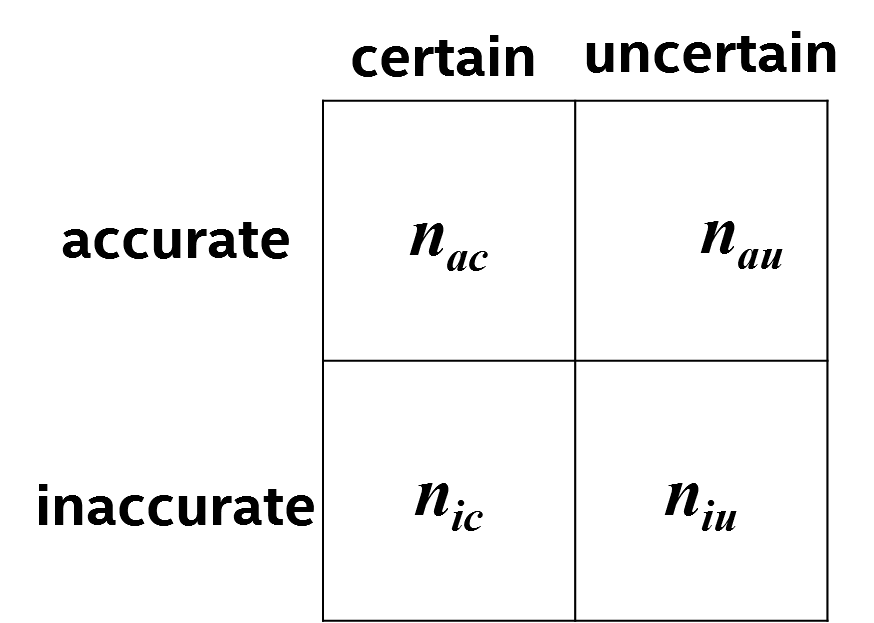}
\caption{ Accuracy vs Uncertainty confusion matrix}
\label{fig:AucVsUnc}
\end{figure}

\begin{figure}[t]
\centering
\includegraphics[width=0.75\linewidth]{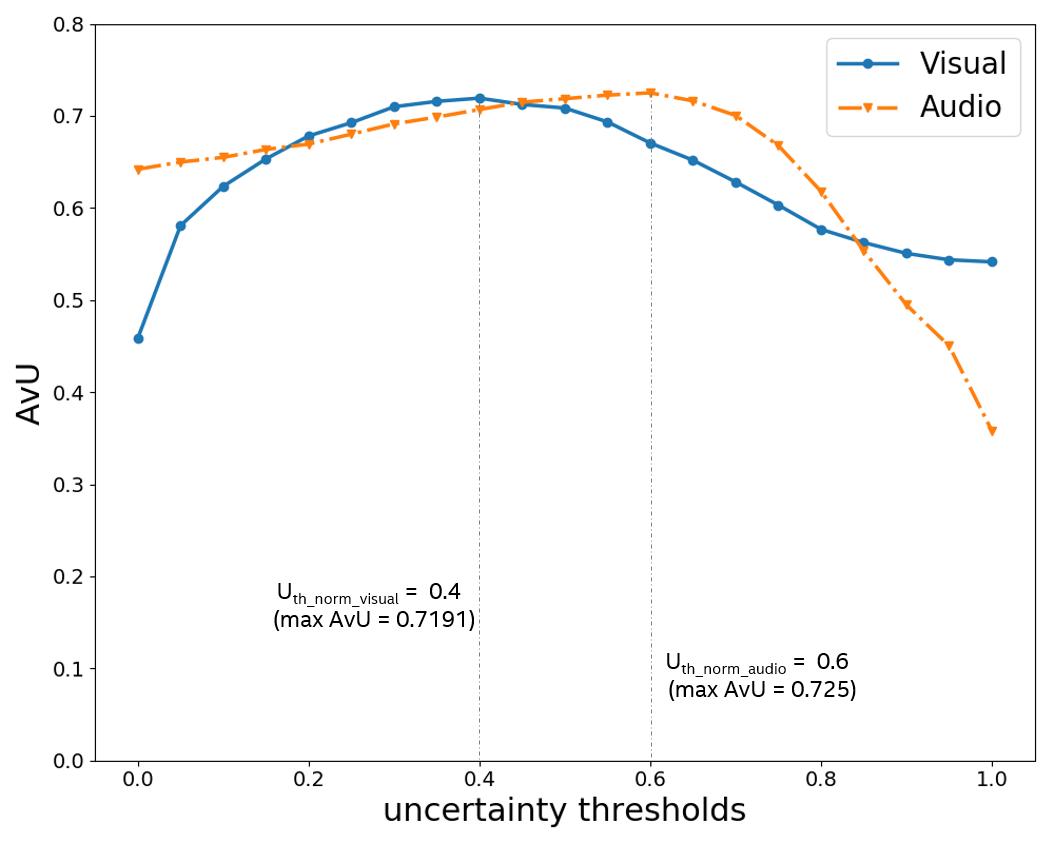}
\caption{ Accuracy vs Uncertainty plots for vision and audio modality. The peak $A\text{v}U$ values represent optimal uncertainty threshold values.}
\label{fig:AucVsUncPlot}
\end{figure}

Figure~\ref{fig:AucVsUnc} shows accuracy vs uncertainty confusion matrix~(proposed in~\cite{mukhoti2018evaluating} for semantic segmentation), which includes number of accurate and certain~($n_{ac}$), inaccurate and uncertain~($n_{iu}$), accurate and uncertain ($n_{au}$), inaccurate and certain($n_{ic}$) predictions. Equation~\ref{eq:auc_vs_unc} provides an accuracy vs uncertainty ($A\text{v}U$) metric obtained from the confusion matrix values.
\begin{equation}
A\text{v}U = \frac{n_{ac} + n_{iu}}{n_{ac} + n_{au} + n_{ic} + n_{iu}}
\label{eq:auc_vs_unc}
\end{equation}
A reliable model will provide higher $A\text{v}U$ score. An uncertainty threshold value that maximizes $A\text{v}U$ metric from individual modalities is the optimal threshold, which is used for multimodal fusion (shown in Figure~\ref{fig:AucVsUncPlot}).
We perform average pooling of the audio-vision predictive distributions if the uncertainty measures are below the optimal threshold values, else we rely on the single modality that has lower uncertainty measure.

For comparison with the non-Bayesian baseline, we maintain the same model depth as the Bayesian DNN model and use three deterministic fully connected final layers for the non-Bayesian DNN model. The dropout is used after every fully connected layer to avoid over-fitting of the model. In the rest of the document, we refer the non-Bayesian DNN model as simply the DNN model.  In the following section, we present the results from our experiments showing the effectiveness of Bayesian DNN over conventional DNN models.

\newcommand{\mysizeb}{0.28}
\begin{figure*}[t]
\centering
\begin{subfigure}[t]{0.33\linewidth}
\captionsetup{
justification=centering}
\centering
\includegraphics[scale=0.23]{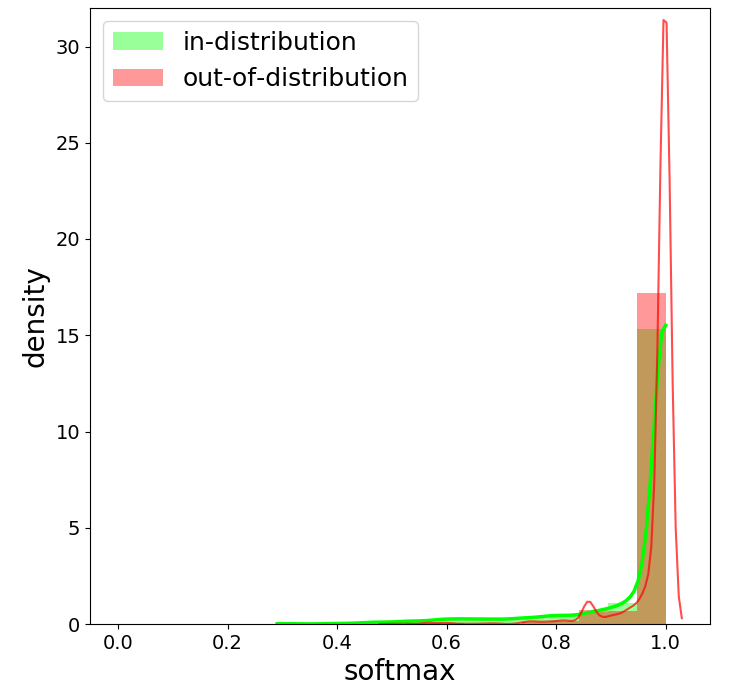}
\caption{DNN\\confidence measure}
\end{subfigure}%
\begin{subfigure}[t]{0.33\textwidth}
\centering
\captionsetup{
justification=centering}
\includegraphics[scale=\mysizeb]{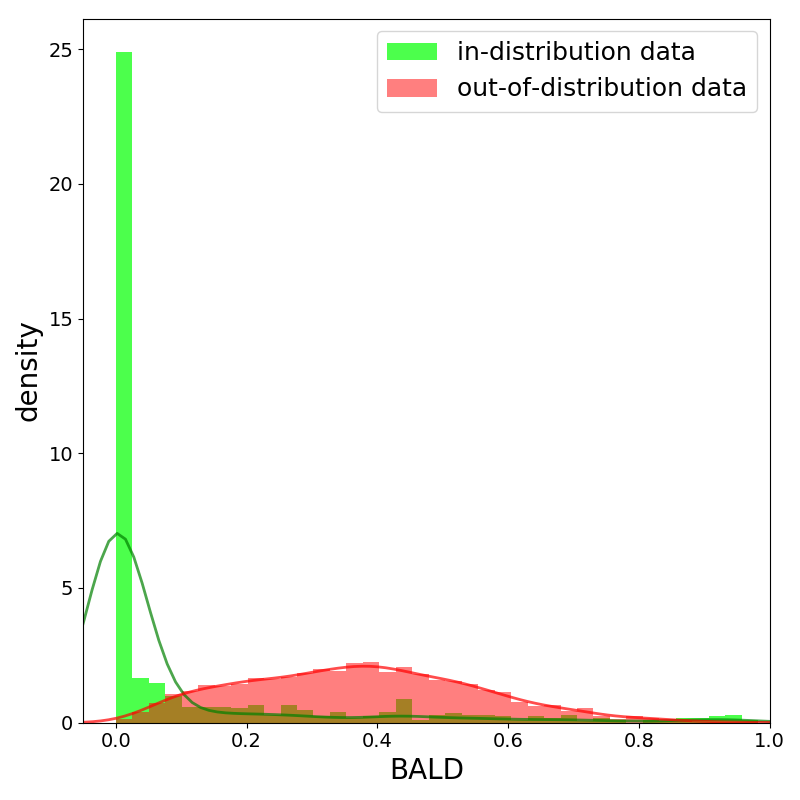}
\caption{Bayesian DNN (MC Dropout) uncertainty measure}
\end{subfigure}
\begin{subfigure}[t]{0.33\textwidth}
\centering
\captionsetup{
justification=centering}
\includegraphics[scale=\mysizeb]{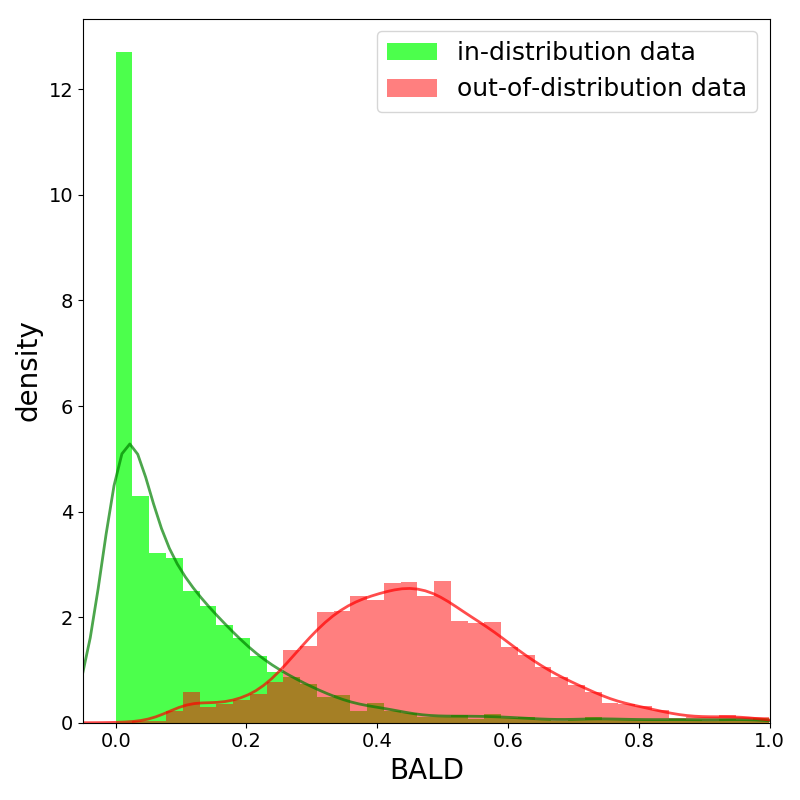}
\caption{Bayesian DNN (Stochastic VI) uncertainty measure}
\end{subfigure}
\caption{\small Density histograms obtained from in- and out-of-distribution samples for the subset of MiT  dataset. (a) DNN confidence measure, (b) Bayesian DNN ()MC Dropout) uncertainty measure and (c) Bayesian DNN (Stochastic VI) uncertainty measure. DNN model indicates high confidence for both the categories(peaked to the right or higher values). Bayesian DNN model uncertainty estimates indicate higher uncertainty for out-of-distribution samples as compared to the in-distribution samples. [The density histogram is a histogram with area normalized to one. Plots are overlaid with kernel density curves for better readability.]}
\label{fig:ConfMeasures_in-out-dist}
\end{figure*}

\section{Results}
\label{sec:results}
We analyze the model performance on the Moments-in-Time (MiT)~\cite{monfort2018moments} dataset. The MiT dataset consists of 339 classes, and each video clip is 3~secs (\textasciitilde90 frames) in length. In this work, we considered a subset of 54 classes as in-distribution and another 54 classes as out of distribution samples. The selected dataset for both the categories include audio information. In order to check whether DNNs can provide a reliable confidence measure, the subset of 54 classes for each category are selected after subjective evaluation to confirm the activities fall into two distinct distribution of classes. 
This will allow the comparison of confidence measures between  DNN and Bayesian DNN models for in- and out-of-distribution classes, and the uncertainty estimates for the Bayesian DNN models (as the DNN model does not provide uncertainty estimates).
\newcommand{\mysize}{0.31}
\begin{figure*}[t]
\centering
\begin{subfigure}[t]{0.33\linewidth}
\centering
\includegraphics[scale=\mysize]{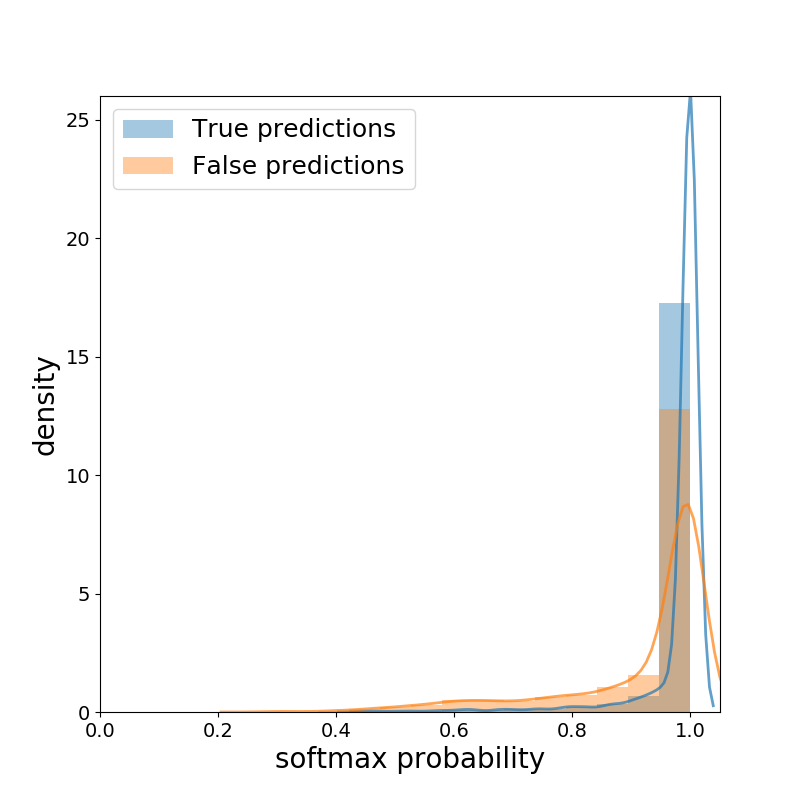}
\caption{\small{DNN model}}
\end{subfigure}%
\begin{subfigure}[t]{0.33\linewidth}
\centering
\includegraphics[scale=\mysize]{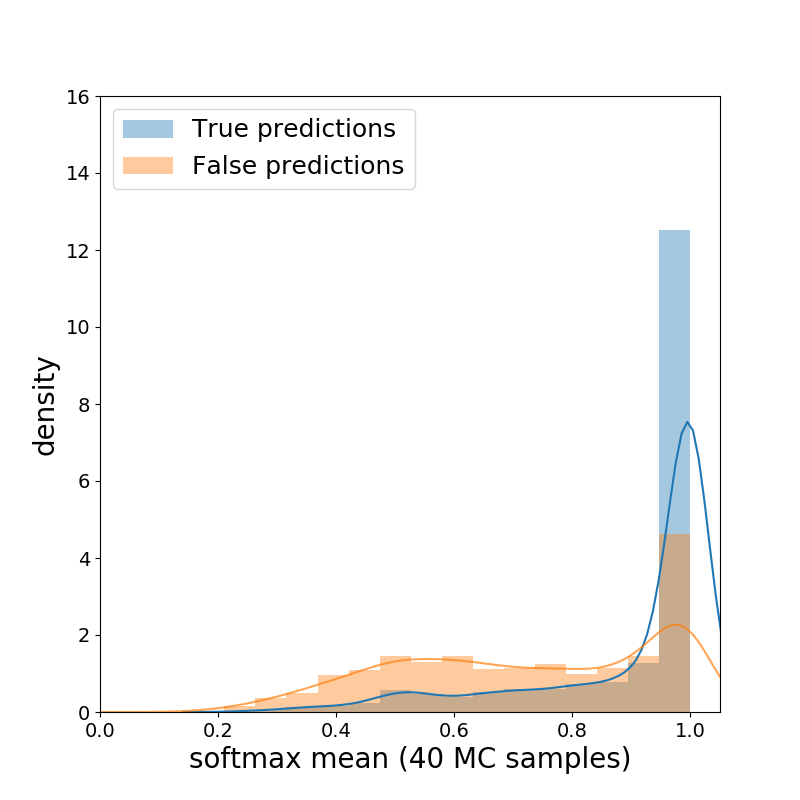}
\caption{\small{Bayesian DNN  (MC Dropout) model}}
\end{subfigure}%
\begin{subfigure}[t]{0.33\textwidth}
\centering
\includegraphics[scale=\mysize]{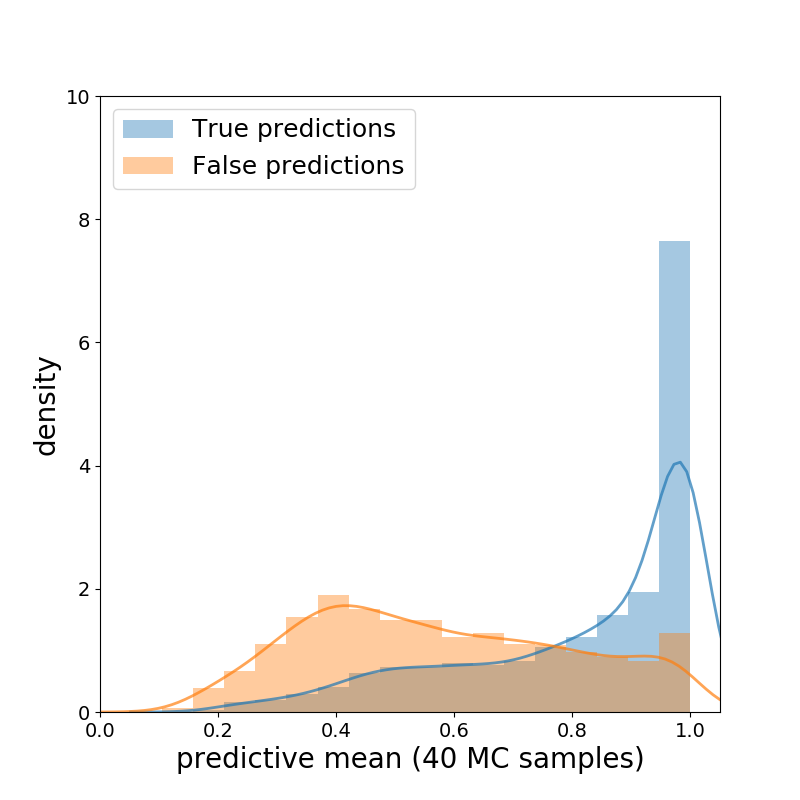}
\caption{\small{Bayesian DNN (Stochastic VI) model}}
\end{subfigure}
\caption{\small{Density histogram of confidence measures for subset of MiT dataset in-distribution true (correct) and false (incorrect) predictions: A distribution skewed towards right (near 1.0 on x-axis) indicates the model has higher confidence in predictions than the distribution skewed towards left. DNN model indicates high confidence for both true and false predictions. Bayesian DNN model shows lower confidence for false predictions while maintaining higher confidence values for the true predictions.  [The density histogram is a histogram with area normalized to one. Plots are overlaid with kernel density curves for better readability.]}}
\label{fig:ConfMeasures}
\end{figure*}

The ResNet-101 C3D DNN model is initialized with pretrained weights for the Kinetics dataset~\cite{kay2017kinetics}. We optimize the model for MiT dataset with transfer learning by training the final fourteen layers. The VGGish model is initialized with pretrained weights for the Audio set~\cite{gemmeke2017audio} dataset. We optimize the model for MiT dataset with transfer learning by training the final five layers. We used stochastic gradient descent (SGD) optimizer with an initial learning rate of 0.0001 and momentum factor of 0.9 along with rate decay when the loss is plateaued.

We trained the ResNet101-C3D vision and VGGish audio architectures using the in-distribution MiT dataset, which includes \textasciitilde150K training and \textasciitilde5.3K validation samples. We select individual vision and audio paths from the model shown in Figure~\ref{fig:VisionBlkDiagram} to obtain single modality results. In the case of Bayesian DNN stochastic~VI model, we perform multiple stochastic forward passes on the final three fully connected variational layers with Monte Carlo sampling on the weight posterior distributions. In our experiments, 40 forward passes provide reliable estimates above which the final results are not affected.  Bayesian DNN model predictive mean is obtained by averaging the confidence estimates from the Monte Carlo sampling predictive distributions.

Bayesian active learning by disagreement (BALD) and predictive entropy uncertainty estimates for Bayesian DNN model are obtained using Equation~\ref{eq:mutual information}~and~\ref{eq:pred_entropy}. Figure~\ref{fig:AucVsUncPlot} shows the accuracy vs uncertainty ($A\text{v}U$) metric plots for audio and vision modalities. An optimal threshold for uncertainty measure that will maximize the $A\text{v}U$ score is computed. For the audiovisual Bayesian DNN results, we perform average pooling of the audio-vision predictive distributions if the uncertainty measures are below the optimal threshold values ($U_{th\_visual}$ and $U_{th\_audio}$), else we fall back to the single modality with lower uncertainty measure. In the case of audiovisual DNN model, average pooling of the softmax confidence values from the two modalities is used.

We compare the proposed stochastic~VI Bayesian DNN with the baseline DNN model. We also compare with well-known Monte Carlo (MC) dropout~\cite{gal2016dropout} approximate Bayesian inference method. For MC dropout, we perform 40 stochastic forward passes with dropout probabilities of 0.5 (same dropout probability is used in the training phase).

\subsection{Uncertainty and confidence measures}

Bayesian DNN models capture uncertainty estimates associated with individual modalities that can be used for multimodal fusion. We compare BALD uncertainty measure (details are in Section~\ref{sec:background}) using in- and out-of-distribution classes from the subset of MiT dataset. Out-of-distribution samples are data points which fall far off from the training data distribution. The DNN models provide softmax probability as the measure of confidence in the results, but do not provide an explicit measure of model uncertainty.

The density histograms for the DNN confidence measure and Bayesian DNN uncertainty measure are plotted in Figure~\ref{fig:ConfMeasures_in-out-dist}. The density histogram is a histogram with area normalized to one. The confidence measure density histogram plots for DNN model (Figure~\ref{fig:ConfMeasures_in-out-dist}~(a)) indicate higher confidence for both in- and out-of-distribution classes. A peak is observed near higher confidence values for out-of-distribution samples indicating incorrect confidence predictions.  The uncertainty estimates obtained from the Bayesian DNN models (Figure~\ref{fig:ConfMeasures_in-out-dist}~(b)~and~(c)) indicate higher uncertainty for the out-of-distribution samples and lower uncertainty values for the in-distribution samples. A peak is observed near higher uncertainty values for out-of-distribution samples indicating reliable predictions.

We compare the confidence measure obtained from the DNN and Bayesian DNN models. The mean of the categorical predictive distribution obtained from Monte Carlo sampling provides the confidence measure for Bayesian DNNs. The confidence measure for the conventional DNN is the softmax probabilities used for the predictions.

The density histograms for the confidence measure are plotted in Figure~\ref{fig:ConfMeasures}. The height (y-axis) of density histogram indicates the distribution of confidence measure. A distribution skewed towards the right (near 1.0 on x-axis) indicates the model has higher confidence in the predictions and the distributions skewed towards left indicate lower confidence.
For true (correct) predictions all three models show confidence measure density histograms peaked near 1.0, indicating reliable predictions. In the case of false (incorrect) predictions, the DNN model still shows confidence measure density histograms peaked near 1.0. On the contrary, Bayesian DNN models show confidence measure density histograms skewed towards lower values indicating more reliable predictions. The proposed stochastic~VI model shows a more pronounced peak towards lower values for false predictions indicating better predictive confidence measure than the MC dropout model.

\begin{figure}[b]
\centering
\includegraphics[width=0.85\linewidth]{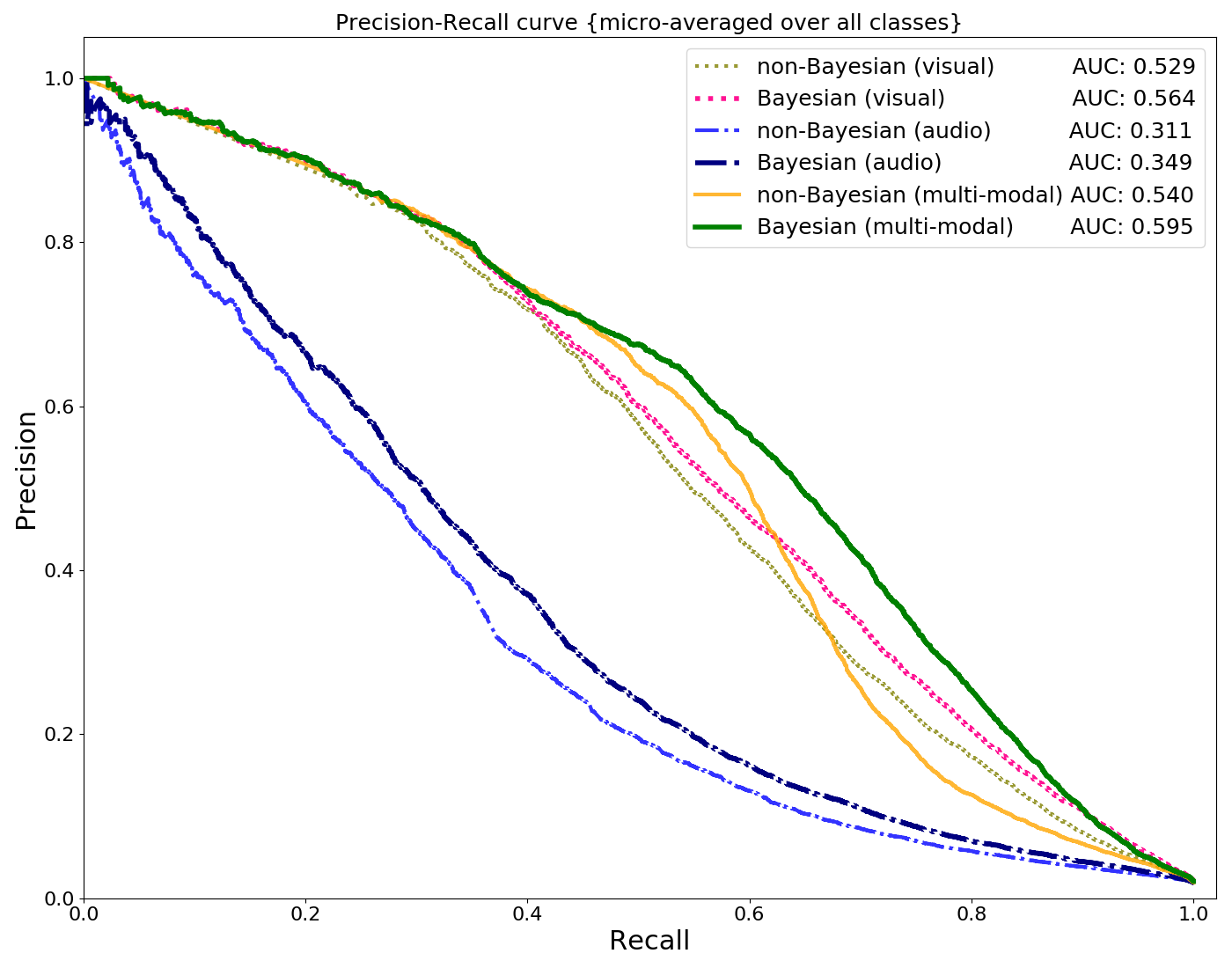}
\includegraphics[width=0.85\linewidth]{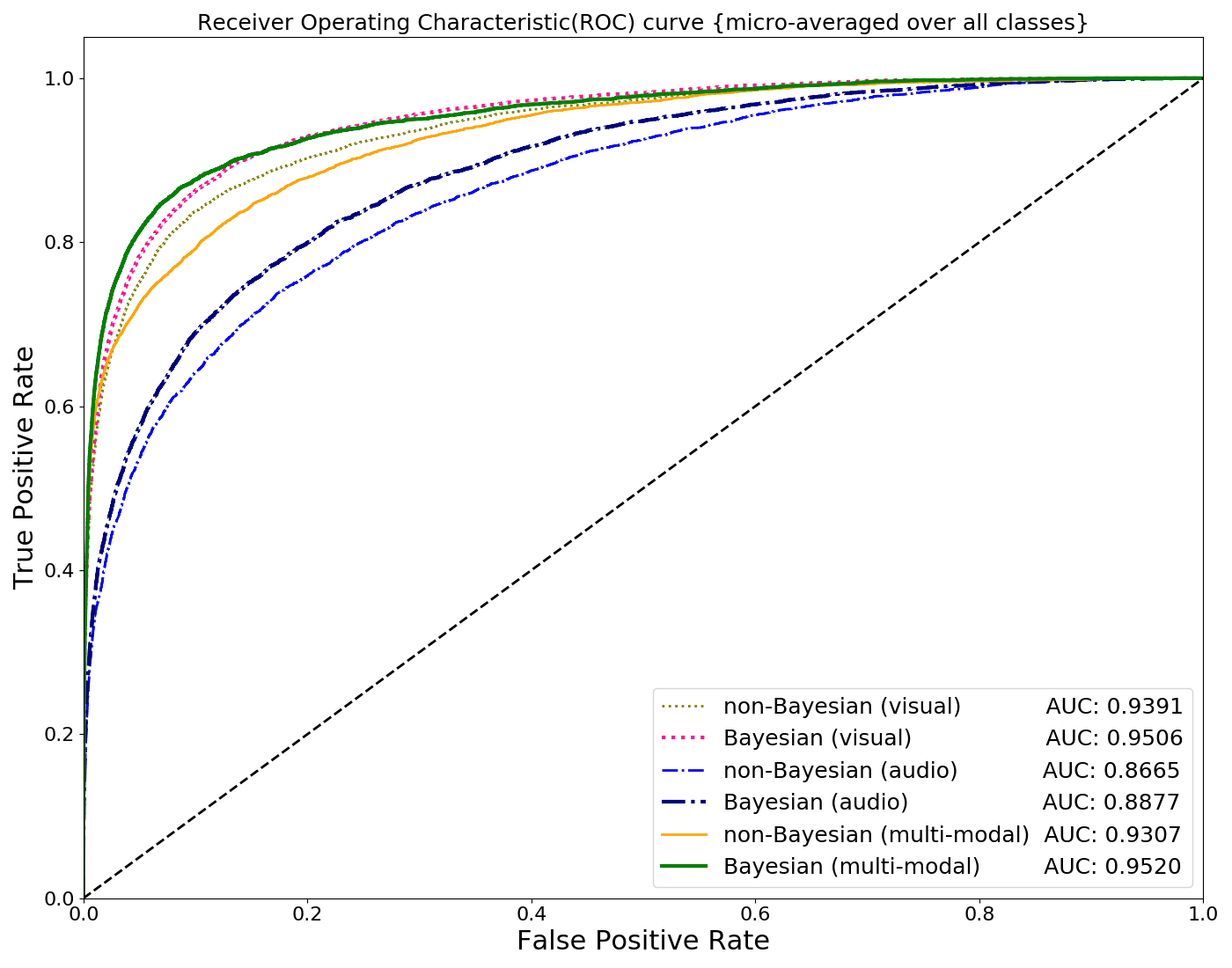}
\caption{ Precision-Recall (top) and ROC (bottom) plots micro-averaged over all the MiT in-distribution classes. }
\label{fig:PrecxRecall Curve}
\end{figure}

\begin{table}[t]
\begin{center}
\begin{tabular}{|l|c|c|}
\hline\hline
\textbf{Model} & \textbf{Top1 (\%)}  & \textbf{Top5 (\%)} \\\hline\hline
\multicolumn{3}{|c|}{\textbf{Vision}} \\\hline
DNN & 52.65 & 79.79 \\\hline
Bayesian DNN (MC Dropout) & 52.88 & 80.10 \\\hline
Bayesian DNN (Stochastic VI) & 53.3 & 81.20 \\\hline\hline
\multicolumn{3}{|c|}{\textbf{Audio}} \\\hline
DNN & 34.13 & 61.68 \\\hline
Bayesian DNN (MC Dropout) & 32.46 & 60.97 \\ \hline
Bayesian DNN (Stochastic VI) & 35.80 & 63.40 \\ \hline\hline
\multicolumn{3}{|c|}{\textbf{Audiovisual}} \\\hline
DNN & 56.61 & 79.39 \\\hline
Bayesian DNN (MC-Dropout) & 55.04 & 80.34 \\ \hline
Bayesian DNN (Stochastic VI) & \textbf{58.2} & \textbf{83.8} \\ \hline
\hline
\end{tabular}
\end{center}
\caption{ Comparison of accuracies for DNN, Bayesian DNN MC Dropout and Stochastic Variational Inference (Stochastic~VI) models applied to subset of MiT dataset (in-distribution classes).}
\label{tab:Accuracy}
\end{table}

\subsection{Model performance comparison}
The classification accuracy for MiT in-distribution samples is presented in Table~\ref{tab:Accuracy}. Bayesian DNN stochastic~VI model consistently provides higher accuracies for individual and combined audio-vision modalities. Bayesian DNN stochastic~VI audiovisual model provides an improvement of 9.2\%~top1 and 3.2\%~top5 accuracies over the Bayesian DNN visual model. Bayesian DNN stochastic~VI model (audiovisual) provides an improvement of 2.8\%~top1 and 5.6\%~top5 accuracies over the baseline DNN model (audiovisual).
The accuracies for Bayesian DNN MC~dropout model are lower than the proposed Bayesian stochastic~VI model.

\begin{figure}[t]
\centering
\includegraphics[width=0.8\linewidth]{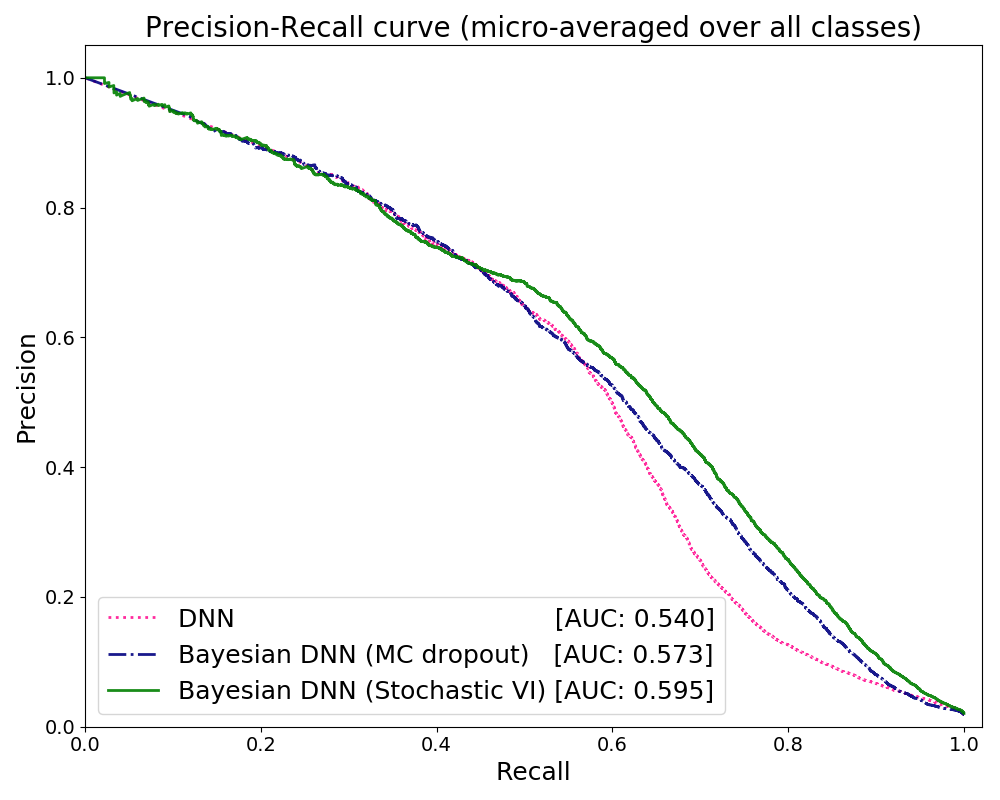}
\includegraphics[width=0.8\linewidth]{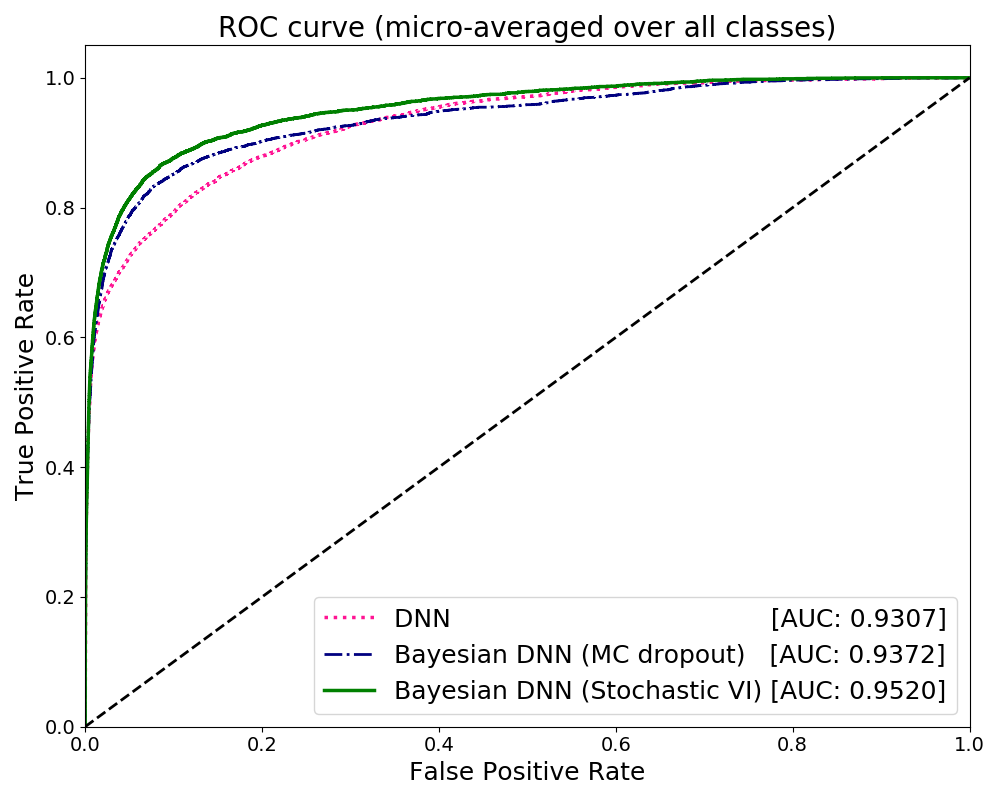}
\caption{\small Precision-recall (top) and ROC (bottom) AUC plots for audiovisual models micro-averaged over all the MiT in-distribution classes.} 
\label{fig:PrecxRecall Curve-mcdroput}
\end{figure}

Figure~\ref{fig:PrecxRecall Curve} shows the comparison of precision-recall and ROC plots using the confidence measures for DNN and Bayesian DNN stochastic~VI models. The proposed model consistently provides higher precision-recall and ROC AUC for individual and combined audio-vision modalities. Figure~\ref{fig:PrecxRecall Curve-mcdroput} shows Bayesian stochastic~VI audiovisual model provides precision-recall AUC improvement of 10.2\%  over the DNN and 3.8\% over the MC Dropout audiovisual models.

\begin{figure}[t]
\begin{center}
\includegraphics[width=0.7\linewidth]{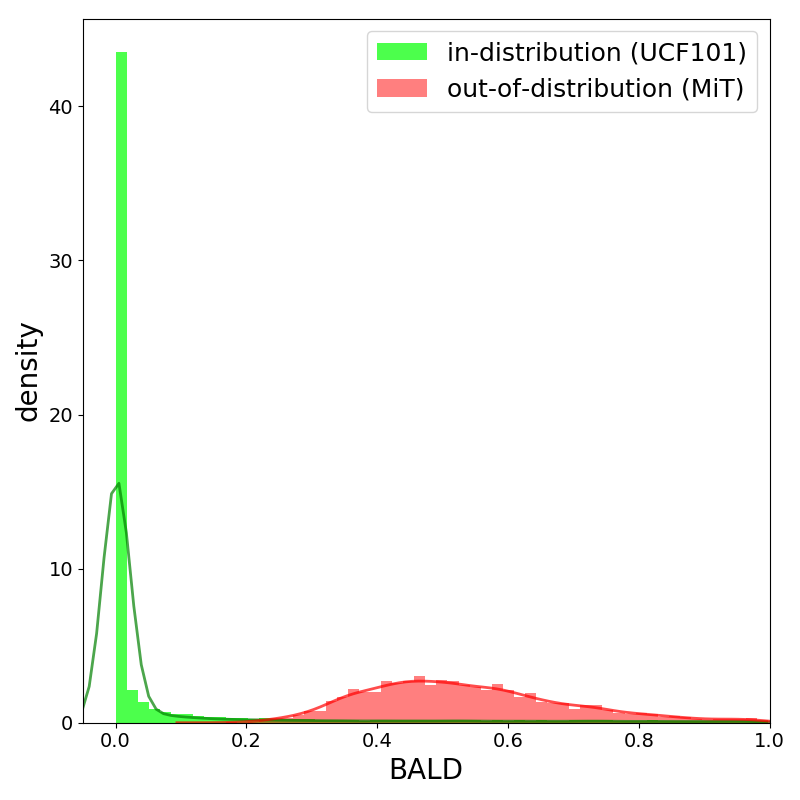}
\end{center}
\begin{center}
\includegraphics[width=0.7\linewidth]{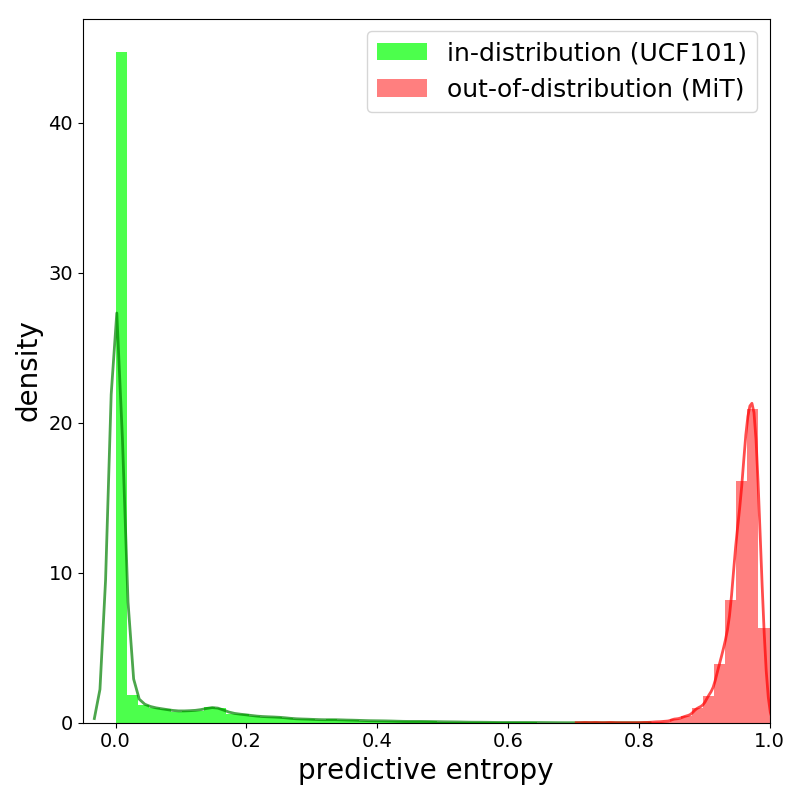}
\end{center}
\caption{ \small Density histogram of uncertainty measures (BALD and predictive entropy) obtained from Bayesian DNN stochastic~VI model. In-distribution samples are from the UCF101 activity recognition dataset and out-of-distribution are from the MiT dataset. The uncertainty measures demonstrate clear separation of in- and out-of-distribution uncertainty distributions. [The density histogram has area normalized to one. Plots are overlaid with kernel density curves for better readability.]}
\label{fig:UncertainityUCF101}
\end{figure}

We also compared the uncertainty estimates obtained from the proposed Bayesian DNN stochastic~VI model using two separate datasets. We compared the UCF101 visual activity recognition dataset, which has 101 activity classes, as in-distribution samples and MiT dataset (vision input) as the out-of-distribution samples. The training of the UCF101 dataset for vision input is done similar to the details provided in Section~\ref{sec:architecture}. The DNN (Top1:~87.5\% and Top5:~97.35\%) and Bayesian DNN (Top1:~88.6\% and Top5:~98.25\%) models provide comparable accuracy values to other results obtained for UCF101 using ResNet-101 C3D model~\cite{hara3dcnns}. The comparison of uncertainty measures for in-distribution and out-of-distribution samples obtained from Bayesian DNN are shown in Figure~\ref{fig:UncertainityUCF101}. Both BALD and predictive entropy (details are in Section~\ref{sec:background}) uncertainty measures indicate a clear separation of uncertainty scores for in- and out-of-distribution samples. 

These results confirm that the proposed Bayesian DNN stochastic~VI model provides reliable confidence measure than the conventional DNN for the audiovisual activity recognition and can identify out-of-distribution samples.

\section{Conclusions}
\label{sec:conclusion}
Effective multimodal activity recognition requires the underlying system to intelligently decide the relative importance of each modality. 
Bayesian inference provides a systematic way to quantify uncertainty in the deep learning model predictions.
Uncertainty estimates obtained from Bayesian DNNs can identify inherent ambiguity in individual modalities, which in turn can benefit multimodal fusion.
In this work, we proposed a novel uncertainty-aware multimodal fusion method using Bayesian DNN architecture that combines deterministic and variational layers.
We evaluate the proposed approach on audiovisual activity recognition using Moments-in-Time dataset.  The results indicate Bayesian DNN can provide more reliable confidence measure compared to the conventional DNNs. The uncertainty estimates obtained from the proposed method have the potential to identify out-of-distribution data. 
The proposed method is scalable to deeper architectures and can be extended to other real-world multimodal applications.

{\small
\bibliographystyle{ieee_fullname}
\bibliography{MAR_ICCV_2019_camera_ready_upload_arxiv}
}

\end{document}